\documentclass[10pt,twocolumn,letterpaper]{article}

\usepackage{iccv}
\usepackage{times}
\usepackage{epsfig}
\usepackage{graphicx}
\usepackage{amsmath}
\usepackage{amssymb}
\usepackage{amsthm}
\usepackage{multirow}
\usepackage{subcaption} 
\usepackage{xspace}
\usepackage{color}
\usepackage{booktabs}
\usepackage{algorithm}
\usepackage{algorithmic}

\usepackage[pagebackref=true,breaklinks=true,letterpaper=true,colorlinks,bookmarks=false]{hyperref}
\newcommand\modelname{DetPTQ\xspace}
 \iccvfinalcopy 

\def\iccvPaperID{4300} 

\ificcvfinal\fi

\begin{document}

\title{Improving Post-Training Quantization on Object Detection with Task Loss-Guided $L_p$ Metric }

\author{Lin Niu $^{\star,\diamond}$\quad Jiawei Liu $^{\star,\diamond}$ \quad Xinggang Wang \quad Wenyu Liu$^{\dagger}$\\
School of EIC, Huazhong University of Science \& Technology, Wuhan, China \\
{\tt\small \{linniu, jiaweiliu, xgwang, liuwy\}@hust.edu.cn}
\and
  Dawei Yang \quad Zhihang Yuan$^{\dagger}$ \\
Houmo AI, Nanjing, China \\
{\tt\small  \{dawei.yang, zhihang.yuan\}@houmo.ai}
}

\maketitle
\ificcvfinal\thispagestyle{empty}\fi

\let\thefootnote\relax\footnotetext{$^\star$ Equal contribution.
$^\diamond$ This work was done when  Lin Niu and  Jiawei Liu were interns at Houmo AI. $^\dagger$ Corresponding authors.}

\begin{abstract}
   Efficient inference for object detection networks is a major challenge on edge devices. Post-Training Quantization (PTQ), which transforms a full-precision model into low bit-width directly, is an effective and convenient approach to reduce model inference complexity. But it suffers severe accuracy drop when applied to complex tasks such as object detection. 
    PTQ optimizes the quantization parameters by different metrics to minimize the perturbation of quantization. 
    The $p$-norm distance of feature maps before and after quantization, $L_p$, is widely used as the metric to evaluate perturbation.
    For the specialty of object detection network, we observe that the parameter $p$ in $L_p$ metric will significantly influence its quantization performance.
    We indicate that using a fixed hyper-parameter $p$ does not achieve optimal quantization performance. 
    To mitigate this problem, we propose a framework, \modelname, to assign different $p$ values for quantizing different layers using an Object Detection Output Loss (ODOL), which represents the task loss of object detection.
    \modelname employs the ODOL-based adaptive $L_p$ metric to select the optimal quantization parameters.
    Experiments show that our \modelname outperforms the state-of-the-art PTQ methods by a significant margin on both 2D and 3D object detectors. 
    For example, we achieve 31.1/31.7(quantization/full-precision) mAP on RetinaNet-ResNet18 with 4-bit weight and 4-bit activation.
\end{abstract}

\section{Introduction}
\label{sec:introduction}
Benefiting from the incredible power of deep learning, object detection networks~\cite{ren2015faster,lin2017focal} have achieved eye-popping performance. However, a large number of parameters and storage requirements present challenges for real-time inference when deployed on edge devices in the real world such as smartphones and electric cars. It is necessary to reduce the memory footprint and computational cost for efficient inference.
\begin{figure*}[t]
  \centering
   \includegraphics[width=0.90\linewidth]{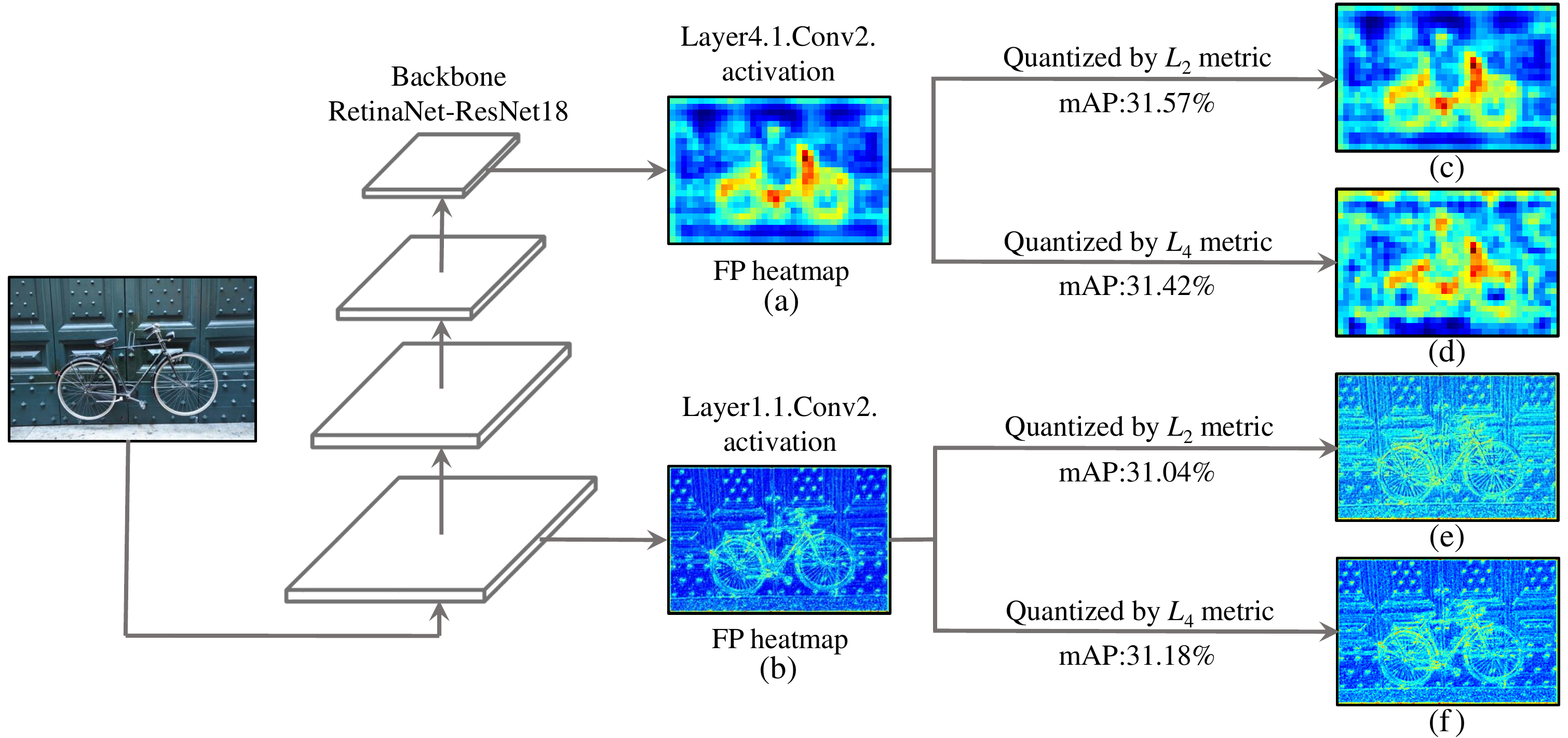}
   \caption{Visualization of the 4-bit PTQ on different activation layers using different $L_p$ metrics in an object detection network (RetinaNet-ResNet18).
All the mAP results are reported on the COCO validation set when quantizing only one activation layer of the network. (c), (d), (e) and (f) show the heatmap of the quantized activation.}
   \label{fig:3.1fig}
\end{figure*}

There are various methods for compressing Convolutional Neural Networks (CNNs) to improve inference efficiency, such as pruning~\cite{he2017channel}, quantization~\cite{gholami2021survey}, and knowledge distillation~\cite{gou2021knowledge}. Among these methods, quantization is particularly promising as it converts full-precision (FP) values to integer grids, reducing computational burden. Quantization can be further divided into two types: Quantization-Aware Training (QAT)\cite{esser2019learned} and Post-Training Quantization (PTQ)\cite{nagel2020up}. Many researchers have retrained object detection networks using labeled training datasets for quantization (i.e., QAT) and achieved great performance. However, QAT can be hindered by the unavailability of labeled datasets due to privacy concerns and the time-consuming training process. PTQ, on the other hand, aims to quantize pre-trained models without retraining, only requiring a small un-labeled calibration dataset. It is popular for deployment in real-world applications. 

Current PTQ methods mainly focus on image classification, with little exploration into their application in object detection. Object detection networks not only classify multiple objects but also regress the location offsets of bounding boxes. These networks have complex architectures, often containing multi-level and multi-dimensional outputs, as opposed to a single vector output in classification networks. These multi-level outputs detect objects of various scales. We observe that different detection levels have different sensitivities to quantization perturbation. When quantizing object detection networks, it suffers a significant performance drop, especially in extremely low-bit settings.


The goal of quantization is to find the optimal quantization parameters that allow the quantized network to achieve the best task performance (minimizing the task loss or maximizing metrics such as mAP~\cite{lin2014microsoft}). However, because labeled data is not available in PTQ, the network's task performance cannot be directly evaluated. To approximate the task loss, previous PTQ methods~\cite{nagel2020up,li2021brecq,wei2022qdrop} often optimize the quantization parameters layer-wise or block-wise to minimize the reconstruction loss for the output of the intermediate layers or blocks, known as local quantization reconstruction. 
They hypothesize that the block-wise (or layer-wise) optimization is mutual-independent and lack attention to the final task output.
Those methods use a fixed metric to evaluate (such as $L_2$ loss) the intermediate feature maps before and after quantization. 
In this paper, we argue that \textit{using a fixed metric without considering its sensitivity to task performance will negatively impact the quantization performance.} 

In this paper, we find that the parameter $p$ in the $L_p$ metric has a significant impact on the quantization of object detection networks. 
We investigate the influence of different $L_p$ metrics on quantization parameters which generated by local quantization reconstruction. We also visualize the quantized feature maps with different $L_p$ metrics.
Taking Figure~\ref{fig:3.1fig} as an example, we observe that setting $L_2$ as quantization metric brings better task performance for the upper level feature map, while setting $L_4$ metric is more beneficial to the lower level feature map. This phenomenon is especially obvious in detection networks. 
We conclude that \textit{different levels of feature maps require different $L_p$ metrics to achieve better detection performance.}

To determine the optimal $p$ values for different levels of features, we propose \modelname to select the $p$ value that makes the local $L_p$ loss approximate to the final task loss. Because labeled data is not available in PTQ, we propose the Object Detection Output Loss (ODOL) which can represent the curve change of the task performance loss when quantizing the network. 
\modelname chooses the $p$ value that makes the $L_p$ loss approximate to the ODOL. To evaluate our method, we experiment on various architectures, including the one-stage 2D detection network RetinaNet~\cite{lin2017focal}, the two-stage 2D detection network Faster RCNN~\cite{ren2015faster}, and the point cloud 3D detection network PointPillar~\cite{Lang_2019_CVPR}. Our work makes the following contributions:
\begin{enumerate}
    \item We indicate that local quantization reconstruction requires an adaptive loss metric for PTQ on object detection networks.
    \item We propose a PTQ framework, \modelname, that assigns different $L_p$ metric to quantize the different layers.
    We propose the ODOL, an approximation of the task loss, to model the global information.
    \item Experiments on 2D and 3D object detectors show that \modelname outperforms the state-of-the-art PTQ methods. For example, \modelname achieves less than 1$\%$ mAP drop with 4-bit PTQ on all of the ResNet-based networks for the first time. 
\end{enumerate}

\section{Related Work}
\subsection{Quantization-aware Training (QAT)}
QAT~\cite{krishnamoorthi2018quantizing,choi2018pact,esser2019learned} uses the entire training datasets to optimize the network for quantization.
The quantization function quantizes tensors in forward-propagation, which has zero-gradient almost everywhere.
QAT methods utilize the Straight Through Estimator (STE)~\cite{bengio2013estimating} to avoid the zero-gradient problem.
Recently, object detection quantization has attracted extensive attention for on-device deployment.
\cite{jacob2018quantization} constrains the bitwidth of objection detection quantization to 8-bit (W8A8), and achieves good results on COCO dataset~\cite{lin2014microsoft}. 
FQN~\cite{li2019fully} quantizes RetinaNet~\cite{lin2017focal} and Faster-RCNN~\cite{ren2015faster} to 4-bit for the first time and demonstrates usable performance. 
AQD~\cite{chen2021aqd} pushes the limits of bit-width in object detection network quantization down to 2-bit. 
Although QAT methods achieve promising results in object detection, extensive training cost and the requirement of labeled datasets prevent its real-world deployment.

\subsection{Post-training Quantization (PTQ)}
Compared with QAT, PTQ~\cite{banner2019post,wu2020easyquant,hubara2021accurate} plays an important role in fast network deployment. 
AdaRound~\cite{nagel2020up} minimizes the $L_2$ loss layer-wisely to optimize a rounding scheme and scale factors.  
Brecq~\cite{li2021brecq} introduces block reconstruction to improve AdaRound. 
QDrop~\cite{wei2022qdrop} shows decent performances at 2-bit by randomly dropping quantized activation during local reconstruction. 
PD-Quant~\cite{liu2022pd} proposes the prediction difference loss (PD-loss) as an approximation of the task loss. 
PD-Quant only explores the metric to classification tasks and requires forward propagation to calculate PD-loss too many times. 
We explore the object-detection-based prediction difference loss.
As the computation cost of object detection is much higher than classification, the adaptive local $L_p$ loss in our \modelname can greatly reduce the quantization overhead. 

Previous works explore the influence of metrics to measure the difference relative to the full-precision for PTQ. 
LAPQ~\cite{nahshan2021loss} find that scale factors from different $L_p$ metrics will cause different quantization errors. It adopts the Powell Algorithm to jointly optimize scale factors from different $L_p$ metrics. 
PTQ4ViT~\cite{yuan2022ptq4vit} proposes to utilize a Hessian-guided metric to evaluate different scale factors for vision transformer.
RAPQ~\cite{yao2022rapq} shows a theory that $p$ value is positively correlated with the sensitivity of the $L_p$ loss value to outliers. It takes the variance information from BN~\cite{ioffe2015batch} to define a BN-based $L_p$ loss.
In our method, we use the global task information (i.e., ODOL) to set the $p$ value. 

Object detection networks often generate multi-level outputs or carry out feature fusion by FPN~\cite{lin2017feature} to detect objects of different scales.
Many PTQ methods such as Brecq and QDrop have successfully extended to objection detection networks but they do not consider the characteristics of object detection networks. 
They use fixed loss in local reconstruction, which is sub-optimal.
Because different levels have different sensitivities, we propose an adaptive loss in local reconstruction.



\section{Preliminaries}
\label{subsec:Preliminaries}
The quantization function transforms a float-point value $x$ to an integer value $\tilde{x}$, and $x^q$ represents the dequantized float value with some error from $x$:

\begin{gather}\label{eq:eq_1}
\tilde{x}= clip(\lfloor \frac{x}{s} \rceil+z,n,m)\\
x^q= (\tilde{x}-z) \cdot s
\end{gather}
where $\lfloor \cdot \rceil$ means that the input $\frac{x}{s}$ is rounded to the nearest integers (i.e.,rounding-to-nearest), which causes the rounding error $\Delta r$.
$clip(\cdot)$ clips the values that lie outside of integer range [$n$, $m$], which causes the clipping error $\Delta c$.
$\tilde{x}$ denotes the integer value, $z$ is zero-point and $s$ denotes the quantization scale factor.
The total perturbation of quantization is $\Delta p= \Delta c+\Delta r = x-x^q$.

PTQ searches for the optimal quantization parameters by solving the problem of minimizing the local reconstruction loss:
\begin{equation}
    \arg\mathop{\min}_{s}\ \|O-O^q\|_2
\end{equation}
where $O$ denotes a FP tensor. $O^q$ is the quantized tensor. $\|\cdot\|_2$ denotes the $L_2$ loss (MSE). For simplisity, we denote all of the quantization parameters with $s$.
Previous PTQ methods~\cite{banner2019post,choukroun2019low} sequentially search the quantization parameters $s$ to minimize the local reconstruction loss layer-by-layer or block-by-block without any training or fine-tuning. This is referred to as the \textbf{simple PTQ method}. 

Recently, some \textbf{advanced PTQ methods} have been proposed, such as AdaRound, BRECQ and QDrop, which reconstruct the weight for better quantization. 
They bring an extra variable $v$ to each weight $w$ and optimize the variable to minimize the local reconstruction loss of the output of a block.
The quantization function and the optimization are formulated as:
\begin{equation}
    \tilde{w}= clip(\lfloor \frac{w+v}{s} \rceil+z,n,m),\;w^q= (\tilde{w}-z) \cdot s
\end{equation}
\begin{equation}
    \arg\mathop{\min}_{V,s}\ \|O -O^q \|_2
\end{equation}
where $V$ is all of the extra variables in a block. 
Our method is orthogonal to the selection of simple PTQ and advanced PTQ, so we denote $s$ as all of the quantization parameters including $V$ for simplicity.



To analyze the influence of the metric for quantization perturbation, we replace the $L_2$ loss with the $L_p$ loss (MSE can be regarded as a special case of $p=2$).
The local reconstruction loss is changed to:
\begin{equation}
    \ \|O-O^q\|_p=(\sum_i \|O_i-O^q_i)\|_p)^{1/p}
\end{equation}
RAPQ~\cite{yao2022rapq} has demonstrated that the $p$ value of the $L_p$ loss is positively correlated with the clipped range. 
For object detection networks, we also find that a larger $p$ for $L_p$ metric leads to a larger scale factor, and vice versa. 
Next, we will deeply explore the influence of different $p$ values.

\section{Method}
\subsection{Influence of Different p Values}
The goal of quantization algorithm is to find the optimal quantization parameters to minimize the performance loss caused by the quantization perturbation. 
PTQ methods usually use MSE ($L_2$) to measure the difference before and after quantization to evaluate the performance loss. 
Since mean Average Precision (mAP) is the main evaluation metric for object detection task, the performance loss can be defined as 
\begin{equation}
    \mathcal{L}_{\text{perf}}=mAP^{fp} - mAP^q
\end{equation}
where $mAP^{fp}$ denotes the performance of the FP network on validation set, which is a constant value for a pre-trained model.
$mAP^q$ is the performance of the quantized network.

To investigate the influence of different $p$ values of $L_p$ metric on different layers, we compare the results of performance loss when quantizing only one layer of the network. 
In Table \ref{tab:tab_1}, we observe that the quantization parameters $s$ are quite different using different $p$.
There are also significant differences in performance losses.
For Layer1.1.Conv2, we observe that the scale factor from $L_3$ has minimal performance loss (0.49), while the scale factor chosen by  $L_2$ (MSE) metric is sub-optimal (performance loss is 0.66). 
As a hyper-parameter of $L_p$ metric, the best $p$ value is different for different layers.
In this example, $L_2$ is the best for Layer4.1.Conv2, while $L_3$ is the best for Layer1.1.Conv2.
Using a fixed $L_p$ metric does not bring to the optimal quantization parameters.
$p$ values need to be adjusted for different layers or blocks.
\begin{table}[t]

\centering
\resizebox{0.85\columnwidth}{!}
{ 
    
\begin{tabular}{crrrr}
    
\toprule
   \multirow{2}{*}{\textbf{Metric}} & \multicolumn{2}{c}{\textbf{Layer1.1.Conv2}} &  \multicolumn{2}{c}{\textbf{Layer4.1.Conv2}}  \\
    \cmidrule(lr){2-3}\cmidrule(lr){4-5}
    & $s$ & $\mathcal{L}_{perf}$ & $s$ & $\mathcal{L}_{perf}$\\
    \midrule
    Min-Max & 0.3472 & 6.00 & 0.5068 & 0.30\\
    $L_1$ & 0.1076 & 2.14 & 0.2246 & 3.17\\
    $L_2$ & 0.1885 & 0.63 & \textbf{0.3462} & \textbf{0.10}\\
    $L_3$ & \textbf{0.2102} & \textbf{0.45} & 0.4273 & 0.21\\
    $L_4$ & 0.2695 & 0.49 & 0.4982 & 0.25\\
      \bottomrule
    \end{tabular} 
    }
    \caption{Comparison of performance loss among different metrics for 4-bit quantization of RetinaNet-Resnet18. We only optimize the scale factor $s$ of one layer's output activation to minimize the local reconstruction error evaluated by $L_p$ metrics. Min-Max is to set the scale factor $s$ that covers all of the activation distribution.}
    \label{tab:tab_1}
\end{table}

The visualization plot of feature maps in Figure~\ref{fig:3.1fig} demonstrates the influence of different $p$ values.
The output activation from Layer1.1.Conv2 is to detect small targets.
From the heatmap (b) using $L_2$, we can observe that the large activation values are clipped to the same value as small activations, causing the detector cannot distinguish the small objects.
A larger $p$ is better to get a larger scale factor $s$, reducing the clipping error on large activations.
From the heatmap (b) using $L_4$, the small objects can be distinguished. 
The results also show quantizing it using $L_4$ (31.18\% mAP) is better than $L_2$ (31.04\% mAP).
Regarding the high-level activation Layer4.1.Conv2, it detects big targets and is more sensitive to global semantic information. The heatmap (c) illustrates that clipping large activation values by $L_4$ metric does not significantly affect the detection of big targets. It is more inclined to use a small $p$ value for the $L_p$ metric to achieve a higher mAP. We conclude that different level activations call for different $L_p$ metrics to achieve accurate detection performance for PTQ. Choosing the right $L_p$ metric to search for the optimal quantization parameters that lead to optimal performance loss is the key issue addressed in the following section.

\subsection{Object Detection Output Loss (ODOL)}

It is a challenge to select the $p$ in $L_p$ metric for local quantization reconstruction.
Note that the goal of quantization algorithm is to find the optimal quantization parameters to minimize the performance loss $\mathcal{L}_{\text{perf}}$.
Because the $p$ value becomes a variable, the formulation of local reconstruction optimization is:
\begin{equation}
    \arg\mathop{\min}_{p}\    \mathcal{L}_{\text{perf}}(s_p^*) 
\end{equation}
\begin{equation}
    s_p^* = \arg\mathop{\min}_{s}\ \|O-O^q\|_p
\end{equation}
where $s_p^*$ is the optimal quantization parameters given $p$ for local reconstruction and $\mathcal{L}_{\text{perf}}(s_p^*)$ is the performance loss with the quantized network using $s_p^*$.
We denote $s^*_{\text{ideal}}$ as the ideal quantization scaling factor, which leads to the lowest performance loss $\mathcal{L}_{\text{perf}}$ assuming we have the labeled data. 
However, there is no label for PTQ's calibration data to calculate the $mAP$ and the $\mathcal{L}_{\text{perf}}$. 
Next, we will propose the Object Detection Output Loss (ODOL) that measures the output difference between the quantized network and FP network as an approximation to $\mathcal{L}_{\text{perf}}$.
We expect that the scale factor, which is optimized by the function, is consistent with $s^*_{\text{ideal}}$. 



The output of object detection networks usually contains two parts: the score for classification and the offset for regression. 
For classification part, we calculate the distance of probabilities predicted by FP network and quantized network for \textbf{all boxes}, which is denoted as class loss $\mathcal{L}_{\text{cls}}$.
For regression part, we calculate the coordinate distance of the boxes predicted by FP network and quantized network for \textbf{positive boxes}, which is denoted as localization loss $\mathcal{L}_{\text{loc}}$.
The function expression of ODOL is formulated as following: 
\begin{align}
\mathcal{L}_{\text{ODOL}}&=\frac{1}{N}\sum_{i=1}^N (\mathcal{L}_{\text{cls,i}}+ \alpha \mathcal{L}_{\text{loc,i}} I_{\text{pos,i}} )\label{eq:eq_5}\\
\mathcal{L}_{\text{cls,i}}&=MSE(c_i,c^q_i)\quad or\quad KL(c_i,c^q_i)\\
\mathcal{L}_{\text{loc,i}}&=L_1(l,l^q)\quad or\quad IOU(l,l^q)
\end{align}
where the $I_{\text{pos,i}}$ is the indicator of whether the $i$-th box is a positive box.
$N$ denotes the number of anchors.
$\alpha$ is a balanced weight.
$c$ and $l$ refer to FP output of scores and coordinates, while $c^q$ and $l^q$ are the output of scores and coordinates from the quantized network~\footnote{ We filter all boxes by a threshold on the score from FP network and select the top 500 boxes. Then we perform NMS~\cite{neubeck2006efficient} on those $k$ boxes to produce positive boxes and indicator $I_{\text{pos,i}}$.
We decode the predicted offset to get the real coordinate for $l$ and $l^q$.
}.


\begin{figure}[tbp]
    \centering
    \subcaptionbox{Faster RCNN-ResNet18\label{faster-rcnn}}
    {
    \includegraphics[width=.48\linewidth]{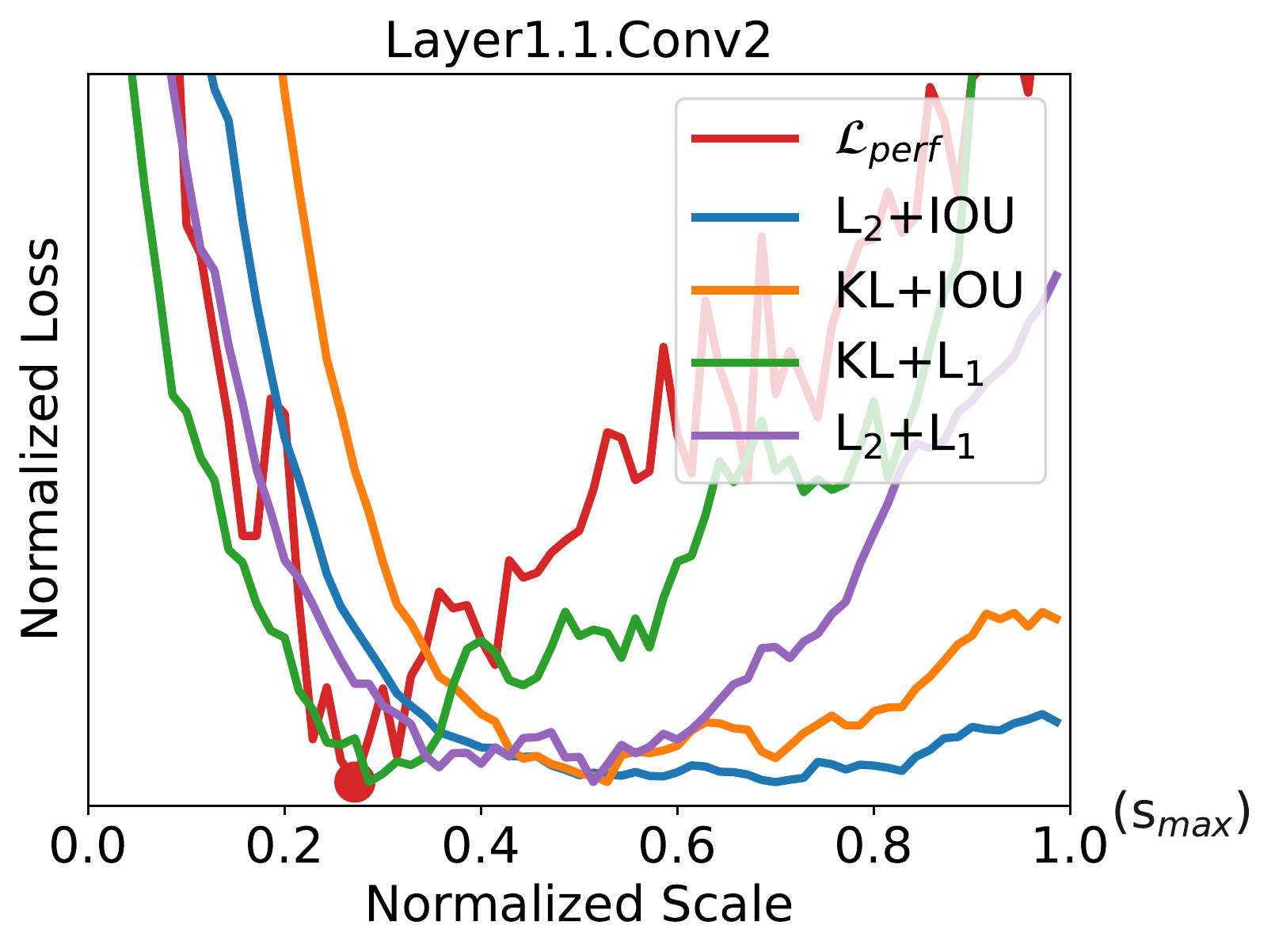}
    \includegraphics[width=.48\linewidth]{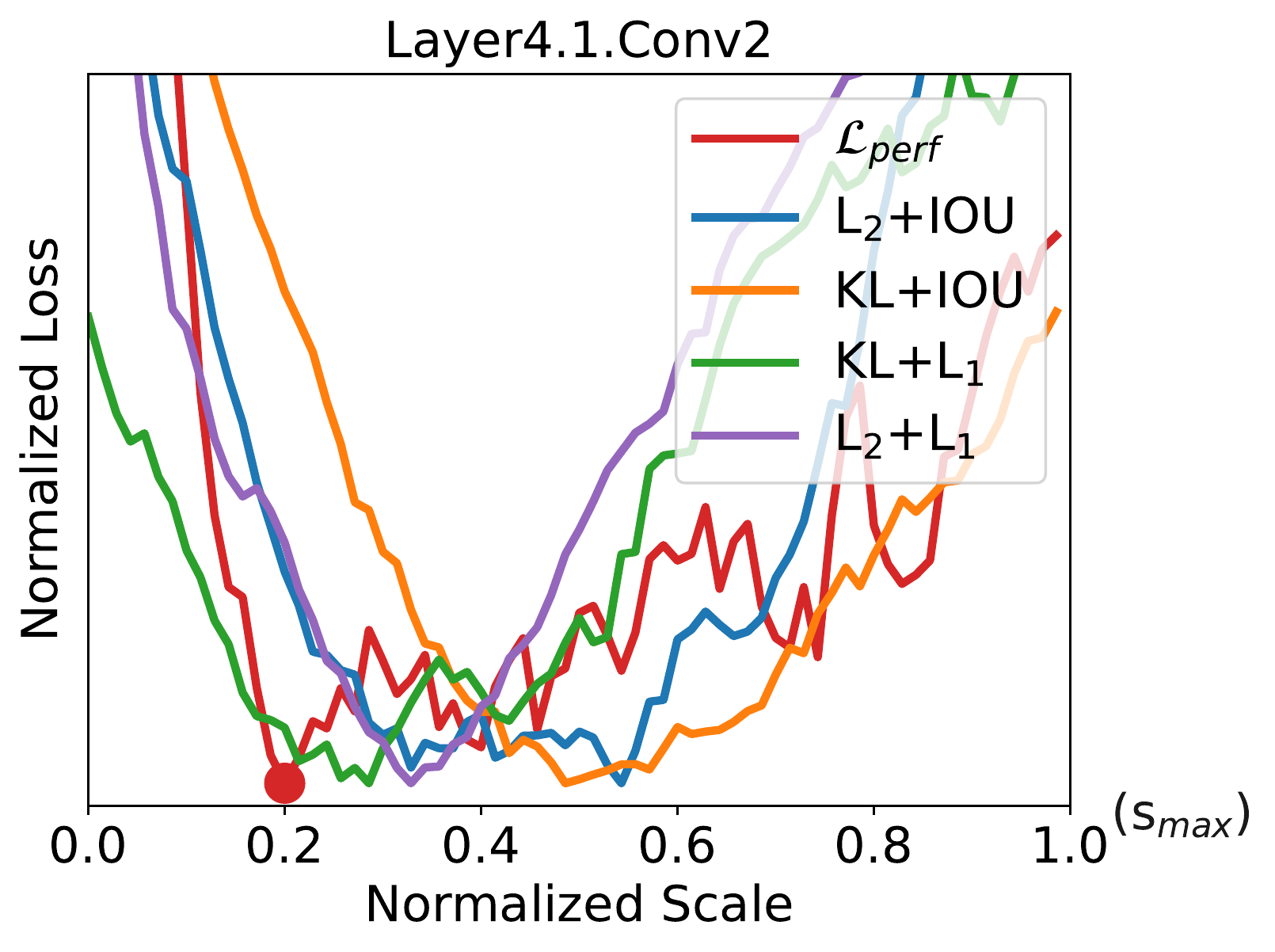}
    }\\
    \subcaptionbox{RetinaNet-ResNet18\label{retinanet}}
    {
    \includegraphics[width=.48\linewidth]{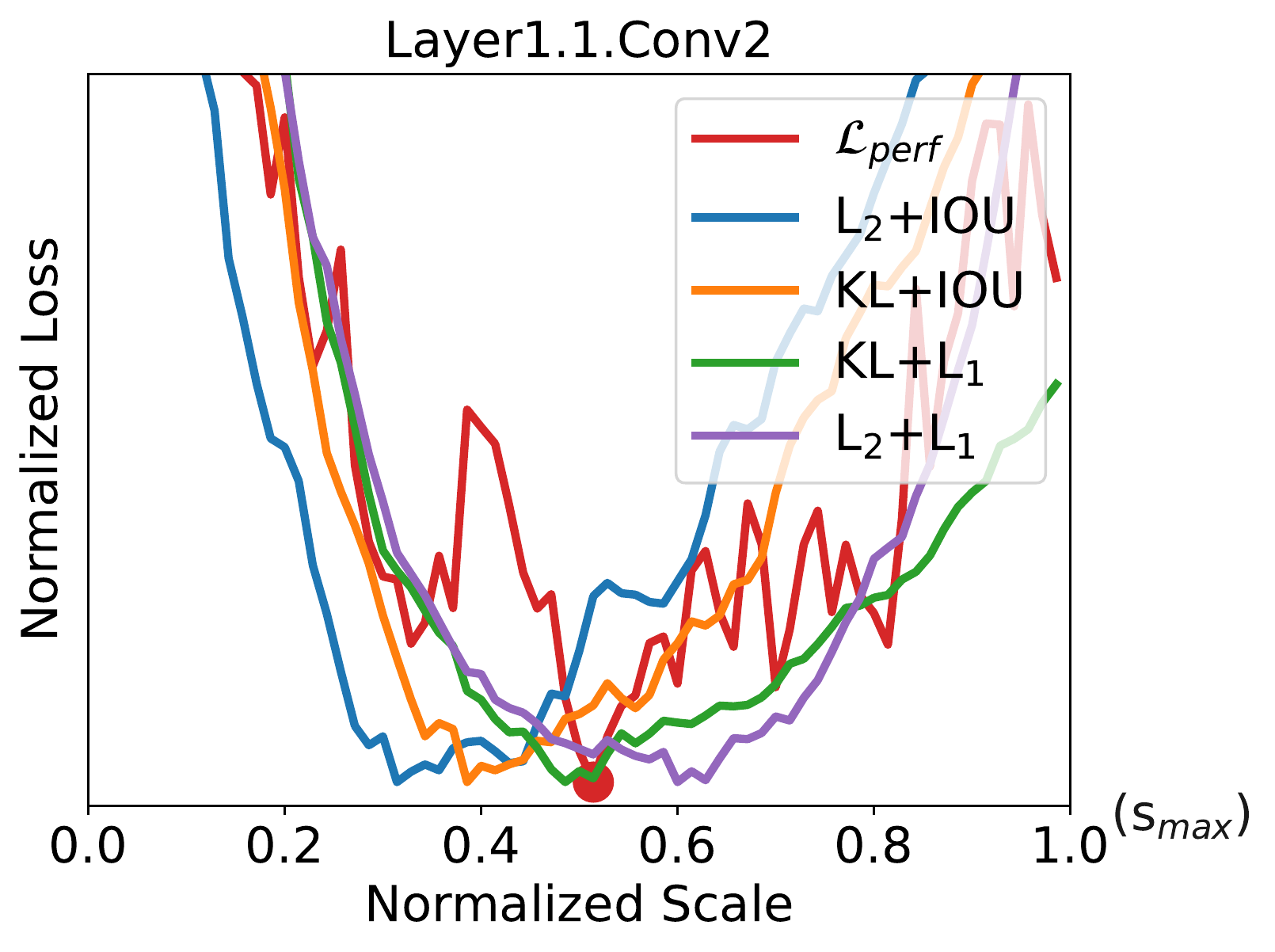}
    \includegraphics[width=.48\linewidth]{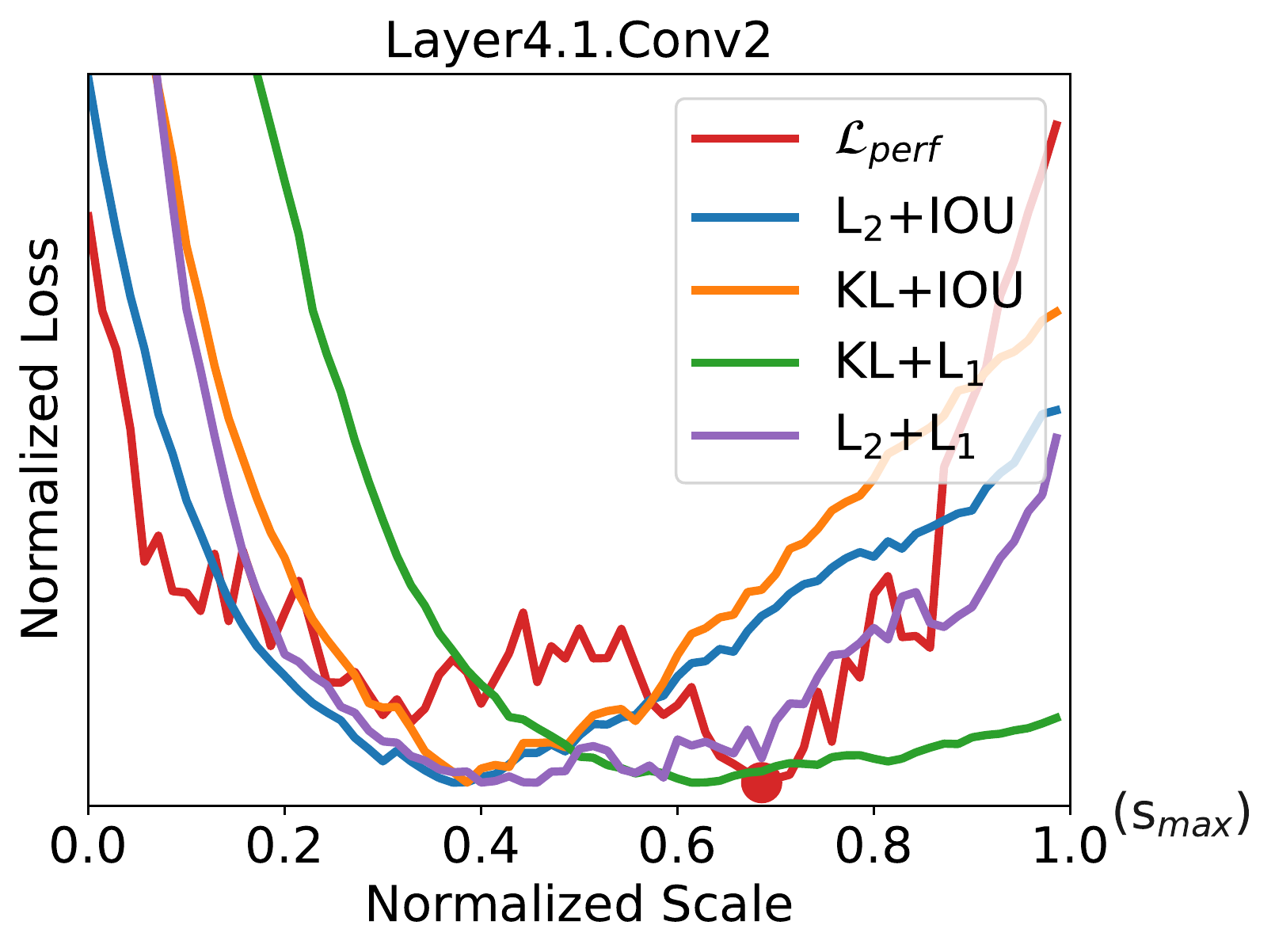}
    }
\caption{Quantization scale-loss curves with different loss functions. 
    The scaling factor $s_{max}$ is calculated by Min-Max quantization and then normalized to 1.0. The loss values (y-axis) are normalized for comparison. 
    }
\label{fig:metric}
\end{figure}

We explore various functions for ODOL. 
We attempt $L_2$ distance and KL distance to measure the classification difference $\mathcal{L}_{\text{cls}}$. 
For localization loss $\mathcal{L}_{\text{loc}}$, we attempt $L_1$ distance and IOU loss~\cite{yu2016unitbox} to measure the distance of coordinate between the FP boxes and quantized boxes. 
To evaluate different loss functions, we compare different scale-loss curves in Figure~\ref{fig:metric}.
The red line is the performance loss, and the ideal scale $s^*_{\text{ideal}}$ is located in the \textcolor{red}{red point}.
We can observe that the scale factor optimized by ODOL with KL+$L_1$ (\textcolor{green}{green line}) is close to $s^*_{\text{ideal}}$. 
Taking an instance, when quantizing the activation of Retinanet-resnet18 Layer1.1.Conv2, $s^*_{\text{ideal}}$ is around 0.48 × $s_{max}$, which is almost the same as ODOL with KL+$L_1$ optimized. 
We will show more experiment results for the different combinations of ODOL in Section \ref{subsec:Ablation}. 
According to the experimental findings, we choose KL as the $\mathcal{L}_{\text{cls}}$ and $L_1$ as the $\mathcal{L}_{\text{loc}}$ as an approximation of $\mathcal{L}_{\text{perf}}$.

\begin{algorithm}[htbp]
    \caption{Network quantization using \modelname}
    \label{alg:algorithm}
    \textbf{Input}: calibration dataset $X^1$ and a network with $L$ blocks.\\  
    \textbf{Output}: quantization parameters of both activation and weight in network
    \begin{algorithmic}[1] 
        \STATE input $X^1$ to FP network to get the FP output;
        \FOR{${B}_{l} = \{B_{i} | i = 1, 2,...L\}$}
            \STATE Initialize a list ${D}$;
            \STATE Input $X^{l}$ from quantized block ${B}_{l-1}$ to FP block ${B}_{l}$;
            
            \FOR{${P} = \{p_{i} | i = 1, 2,...k\}$}
                \STATE Optimize only activation quantization parameters to minimize Eq~\ref{eq:local_reconstruction_ODOL} in block ${B}_{l}$ to get $s_p^*$;
                 \label{line6}
                \STATE Quantize activations in ${B}_{l}$ to get $O^q$;
                \STATE Input $O^q$ to the following FP network to get output of partial-quantized network;
                \STATE Calculate $\mathcal{L}_{\text{ODOL}}(s_p^*)$ using Eq~\ref{eq:eq_5};
                \STATE Append $\mathcal{L}_{\text{ODOL}}(s_p^*)$ to list $D$;
            \ENDFOR
            \STATE $index \leftarrow argmin(D)$;
            \STATE $p^* \leftarrow P[index]$;
            \STATE Optimize quantization parameters for both activation and weight to minimize Eq~\ref{eq:local_reconstruction_ODOL} using $p^*$ in block ${B}_{l}$;
             \label{line14}
            \STATE Quantize weights and activations in ${B}_{l}$ to get $O^q$;
            \STATE $O^q$ as the input $X^{l+1}$ to next FP block ${B}_{l+1}$;
        \ENDFOR
    \end{algorithmic}
    \label{alg:alg_1}
\end{algorithm}
\begin{table*}[htb!]

    \centering
    \resizebox{0.90\textwidth}{!}
    { 
    \begin{tabular}{lrrrrrrr}
    \toprule
   \multirow{2}{*}{\textbf{Method}} & \multirow{2}{*}{\textbf{Bits(W/A)}} & \multicolumn{3}{c}{\textbf{Faster RCNN}} &  \multicolumn{3}{c}{\textbf{RetinaNet}}  \\
    \cmidrule(lr){3-5}\cmidrule(lr){6-8}
    & & ResNet-18 & ResNet-50 & MobileNetV2 & ResNet-18 & ResNet-50 & MobileNetV2\\
    \midrule
    FP & 32/32 & 34.23 & 38.46  & 29.91 & 31.67 & 37.38 & 28.40 \\
    \midrule
    MSE& 4/8 &31.02 &35.35&19.02&28.64 &34.88&9.85                          \\
    Cosine& 4/8 &30.61&34.49&14.41&28.43 &34.42&2.24                       \\
    BN-based adaptive $L_p$& 4/8 &30.57&33.27&18.76&27.21&35.04&10.01                     \\
    DetPTQ* (Ours) & 4/8&\textbf{32.96}&\textbf{37.04}&\textbf{19.96}&\textbf{ 28.95}&\textbf{36.51}&\textbf{12.42}             \\
     \midrule
    MSE& 4/4 &27.75&29.10&8.04&24.81&30.31&2.74                       \\
    Cosine&4/4 &21.59&19.77&0.27&21.48&18.97&0.04                      \\
    BN-based adaptive $L_p$& 4/4 &24.77&28.25&8.12&25.23 &27.43&3.38                     \\
    DetPTQ* (Ours) & 4/4 &\textbf{30.04}&\textbf{31.18}&\textbf{9.87}&\textbf{ 27.67}&\textbf{32.13}&\textbf{5.65}                    \\
      \bottomrule
    \end{tabular} 
    }
    \caption{Comparison DetPTQ* with the simple PTQ method optimized by various metrics on COCO. W/A means the bit which weight/activation is quantized to.}
    \label{tab:table22}
\end{table*}
\subsection{\modelname Framework}
One intuition is that we can use ODOL as a metric to directly optimize the quantization parameters to minimize 
 the distance for the intermediate layer’s activation before and after quantization, just as PD-Quant~\cite{liu2022pd} and NWQ~\cite{nwq} do. However the computation cost of object detection is very large. Repeated forward propagation to calculate ODOL is not realistic.
In the section above, we observe that the parameter $p$ in $L_p$ metric has a significant influence on quantization.
And we propose ODOL as an approximation of performance loss to determine $p$ value for $L_p$ metric.
In this case, ODOL is computed as many times as the number of candidate $p$ values.
The local quantization reconstruction using ODOL can be formulated as:
\begin{equation}
    \arg\mathop{\min}_{p}\   \mathcal{L}_{\text{ODOL}}(s_p^*) 
\end{equation}
\begin{equation}
s_p^*  = \arg\mathop{\min}_{s}\ \|O-O^q\|_p
\label{eq:local_reconstruction_ODOL}
\end{equation}
Next, we will propose our PTQ framework, \modelname.
The same as previous methods, we quantize blocks in sequential order so that the perturbation accumulated in earlier blocks can be made up. 
When quantizing the $l_{th}$ Block (${B}_{l}$), the blocks before ${B}_{l}$ have been quantized and the blocks after ${B}_{l}$ are in floating-point. 

As shown in Algorithm~\ref{alg:alg_1}, the quantization process for a block includes two steps.
The first step is to select the $p$ value. 
Our experiments in Section~\ref{subsec:Ablation} shows the quantization of weight is not sensitive to the selection of $p$.
Therefore, we only use the scale factor for activation values to speedup the search of $p$.
Given a set of values ${P}$ = \{$p_{i}$ $|$ $i = 1, 2,...k$\}for $L_p$ metric, we can optimize the inner problem in Eq~\ref{eq:local_reconstruction_ODOL} to get a set of activation scale factors ${S}$ = \{$s_{p_i}^*$ $|$ $i = 1, 2,...k$\} by minimizing the given $L_p$ metric~\footnote{We set $P=\{1,1.5,2,2.5,3,3.5,4,4.5\}$ in our experiments.}. 
Because there are only a small number activation scale factors in a block, the $p$ selection process executes quickly.
The second step is to optimize to the quantization parameters for both activation and weight to minimize the local reconstruction loss using the selected $p$ value.
The output of the quantized block is used as the input of the next layer.

\begin{table*}[htb!]

    \centering
    \resizebox{0.87\textwidth}{!}
    { 
    
    \begin{tabular}{lrrrrrrr}
    
    \toprule
   \multirow{2}{*}{\textbf{Method}} & \multirow{2}{*}{\textbf{Bits(W/A)}} & \multicolumn{3}{c}{\textbf{Faster RCNN}} &  \multicolumn{3}{c}{\textbf{RetinaNet}}  \\
    \cmidrule(lr){3-5}\cmidrule(lr){6-8}
    & & ResNet-18 & ResNet-50 & MobileNetV2 & ResNet-18 & ResNet-50 & MobileNetV2\\
    \midrule
    FP & 32/32 & 34.23 & 38.46  & 29.91 & 31.67 & 37.38 & 28.40 \\
     \midrule
    RAPQ~\cite{yao2022rapq} & 4/4 & 31.17 & 32.76 & 20.96  & 29.03 & 33.72 & 21.76 \\
    AdaQuant~\cite{hubara2021accurate} & 4/4 & 32.89 & 35.03 & 22.77  & 30.34 & 35.84 & 22.27 \\
    AdaRound~\cite{nagel2020up} & 4/4 & 32.45 & 34.22 & 23.32  & 30.41 & 35.59 & 22.16 \\
    BRECQ~\cite{li2021brecq} & 4/4 & 32.47 & 34.31 & 23.76  & 30.42 & 35.71 & 22.23 \\
    QDROP~\cite{wei2022qdrop} & 4/4 & 33.13 & 36.98 & 25.04  & 30.54 & 36.13 & 25.71 \\
    \modelname (Ours) & 4/4 & \textbf{33.82} & \textbf{37.82} & \textbf{25.49}  & \textbf{31.12} & \textbf{36.83} & \textbf{27.14}  \\
    \midrule
    AdaRound & 3/3 & 29.97 & 32.94 & 18.56  & 27.88 & 32.36 & 15.87\\
    BRECQ & 3/3 & 30.12 & 33.32 & 19.02  & 28.07 & 32.84 & 17.02 \\
    QDROP & 3/3 & 30.93 & 34.32 & 21.31  & 28.48 & 33.45 & 18.73\\
    \modelname (Ours) & 3/3 & \textbf{32.11} & \textbf{34.97} & \textbf{21.94}  & \textbf{29.30} & \textbf{34.48} & \textbf{19.40} \\
    \midrule
    AdaRound & 2/4 & 28.44 & 31.25 & 19.03  & 24.73& 32.87 & 15.81 \\
    BRECQ & 2/4 & 29.21 & 32.10 & 19.45  & 27.76 & 33.11 & 16.02 \\
    QDROP & 2/4 & 30.68 & 34.41 & \textbf{21.56}  & 28.51 & 33.52& 18.62 \\
    \modelname (Ours) & 2/4 & \textbf{30.72} & \textbf{35.31} & 21.31  & \textbf{28.93} & \textbf{34.87}& \textbf{19.35}  \\
      \bottomrule
    \end{tabular} 
    }
    \caption{Comparison \modelname with various SOTA advanced PTQ methods on COCO. }
    \label{tab:tab_2}
\end{table*}
\begin{table*}[htb!]

    \centering
    \resizebox{0.90\textwidth}{!}
    { 
    
    \begin{tabular}{lrrrrrrr}
    
    \toprule
   \multirow{2}{*}{\textbf{Method}} & \multirow{2}{*}{\textbf{Bits(W/A)}} & \multicolumn{3}{c}{\textbf{PointPillar 3D-Box}} &  \multicolumn{3}{c}{\textbf{PointPillar Bounding-Box}}  \\
    \cmidrule(lr){3-5}\cmidrule(lr){6-8}
    & & easy@AP40 & moderate@AP40 & hard@AP40 &easy@AP40 & moderate@AP40 & hard@AP40\\
    \midrule
    FP & 32/32 & 76.72 & 64.48 & 60.79 & 84.05 & 75.44 & 72.58\\
    \midrule
    BRECQ~\cite{li2021brecq} & 4/8& 72.93 & 60.78 & 58.11 & 81.23 & 73.21 & 70.76\\
    QDROP~\cite{wei2022qdrop} & 4/8 & 74.77 & 62.04 & 58.81 & 82.80 & 73.96 & 71.43\\
    \modelname (Ours) & 4/8 & \textbf{76.02} & \textbf{63.89} & \textbf{59.92} & \textbf{83.45} & \textbf{74.66} & \textbf{72.04}\\
    \midrule
    BRECQ & 4/4& 62.83 & 51.24 & 48.03 & 77.45 & 68.26 & 66.01\\
    QDROP & 4/4 & 64.42 & 52.63 & 49.68 & 79.51 & 70.52 & 67.06\\
    \modelname (Ours) & 4/4 & \textbf{68.03} & \textbf{55.85} & \textbf{52.79} & \textbf{83.02} & \textbf{73.27} & \textbf{70.42}\\
      \bottomrule
    \end{tabular} 
    }
    \caption{Comparison on PointPillar with various PTQ algorithms on KITTI.}
    \label{tab:table_3D}
\end{table*}
\section{Experimental Results}
\label{sec:results}

To evaluate the effectiveness of our proposed \modelname, we experiment on the COCO~\cite{lin2014microsoft} benchmark for 2D object detection and KITTI~\cite{Geiger2013IJRR} benchmark for 3D object detection. 
We compare \modelname  with the existing PTQ approaches on various CNN detectors, including RetinaNet~\cite{lin2017focal}, FasterRCNN~\cite{ren2015faster} and PointPillar~\cite{Lang_2019_CVPR}. 
Performance of 2D detection is evaluated by standard COCO metrics with mean average precision (mAP) on the validation set. 
As to KITTI, we pick the standard metrics and report the 3D box and 2D bounding box results at AP40.

\subsection{Implementation Details}

All the FP models in our paper use open-source codes from MMDetection~\cite{mmdetection} and MMDetection3D~\cite{mmdet3d2020}.
For COCO, we randomly pick a total of 256 training samples with a shorter edge to 800/600 pixels for ResNet~\cite{he2016deep} / MobileNetV2~\cite{sandler2018mobilenetv2} as the calibration dataset. As to KITTI, the number of calibration samples is set to 128. 
Following the implementation previous work~\cite{li2021brecq,wei2022qdrop}, the head of the network keeps full-precision, the first and the last layer are quantized to 8 bits.
We execute block reconstruction for backbone and layer reconstruction for neck, respectively. 
We execute all experiments on Nvidia Tesla V100 GPU.
 
The balancing hyper-parameter $\alpha$ in Equation \ref{eq:eq_5} is 0.1/ 0.001 for $L_1$/IOU in the exploration of ODOL.  
To obtain the FP \textbf{positive boxes} in ODOL, the score threshold $\theta$ is 0.05 and the NMS threshold is 0.5. 
Given a $p$ value for $L_p$, we optimize all activation scale factors in a block by Adam optimizer with the learning rate 3e-4, iterations 5000 in Algorithm \ref{alg:alg_1} line \ref{line6}. 
We also optimize the scaling factors of convolution layers which do not belong to any block using Adam optimizer.
The setting of local quantization reconstruction (i.e., Algorithm \ref{alg:alg_1} line \ref{line14}) is kept the same with QDrop, except that the warmup of rounding loss is 0.4.

\subsection{Comparison with other PTQ Methods}
As described in Section \ref{subsec:Preliminaries}, PTQ can be divided into simple PTQ and advanced PTQ by whether optimizing the rounding variable. Our method can be applied to both simple and advanced PTQ. We try our ODOL-based adaptive $L_p$ metric on the simple PTQ method and denote it as DetPTQ*, while we adopt DetPTQ in advanced PTQ to get higher quantization accuracy.

\subsubsection{Result on COCO Benchmark}
For simple PTQ, we compare the standard methods that grid searches the quantization parameters by minimizing the MSE~\cite{nagel2021white} and Cosine distance~\cite{wu2020easyquant} between activations before and after quantization.
We also compare a related work~\cite{nahshan2021loss}, which uses the variance information from BN~\cite{ioffe2015batch} to define a BN-based $L_p$ metric to conduct the local block-wise optimization. 
We denote this method as BN-based adaptive $L_p$.


According to the findings presented in Table \ref{tab:table22}, it is evident that the implementation of DetPTQ* can lead to a substantial enhancement in the simple PTQ method. 
It should be noted that relying solely on local information may not always yield accurate results. 
For instance, MSE, Cosine and BN-based adaptive $L_p$ obtain 27.75, 21.59 and 24.77 mAP on W4A4 Faster RCNN-ResNet18, while DetPTQ* exhibits an improvement of 30.04 mAP for the simple PTQ method. 
Our DetPTQ* demonstrates remarkable improvement in the performance across all networks and bitwidth settings. 
This outcome highlights the efficacy of utilizing global information from the performance loss to derive a better quantization performance.

We also compare our method with strong advanced baselines including AdaRound, BRECQ and QDROP~\footnote{As certain models are unavailable, we present the results of other quantization methods utilizing their open-source codes with all FP models from MMDetection. It is possible that our values may differ from those reported in their papers.}. 
The results are shown in Table \ref{tab:tab_2}.
With W4A4 quantization, we observe that \modelname  achieves less than 1$\%$ mAP drop on the ResNet-based  object detection networks.
As for W3A3 quantization, the task becomes harder but our method can also achieve significant improvement.
In the most difficult W2A4 setting, our method can still improve the performance of the model in most cases.
For instance, \modelname  achieves 1.42$\%$ mAP improvement on RetinaNet-ResNet50 compared to the baseline QDrop. Taking all experiments together, our method is orthogonal to the selection of simple and advanced PTQ as well as provides a convincing uplift on the object detection network.

\subsubsection{Result on KITTI Benchmark}
We choose PointPillar with SECOND backbone to evaluate the performance of \modelname on the KITTI benchmark for 3D object detection. For PointPillar, there is an additional head known as the direction prediction head. In addition to the localization offset, we decode the direction output to calculate the real coordinate, and then execute the localization loss ($\mathcal{L}_{loc}$) to measure the coordinate distance of \textbf{positive boxes} between the FP and quantized outputs.

As Table \ref{tab:table_3D} shows, the results of our \modelname are indeed close to FP model with negligible accuracy drop on PointPillar-SECOND detector. At W4A8, we achieve a decrease of less than 1\% compared to FP accuracy for the first time.
As for the challenging W4A4, our \modelname achieves $0.61\%\sim3.61\%$ improvement compared with strong baselines.
For 3D box @AP40, we outperform QDrop by a large margin of 3.61\%/3.22\%/3.11\% for easy/moderate/hard. 
\subsection{Ablation Study}   
\label{subsec:Ablation}

\begin{table}[t]

    \centering
    \resizebox{0.85\columnwidth}{!}
    { 
    
    \begin{tabular}{crrr}
    
    \toprule
   \textbf{Model} & \textbf{Variable} & \textbf{Optimal $p$} & \textbf{\#Params}\\
    \midrule
    \multirow{2}{*}{Layer1.1.Conv2} & $s_a$ & 3.5 & 2\\
     & $s_a + V$ & 3.5 & 2+1152\\
    \midrule
    \multirow{2}{*}{Layer4.1.Conv2} & $s_a$ & 1.5 & 2\\
     & $s_a + V$ & 1.5 & 2+9216\\
      \bottomrule
    \end{tabular} 
    }
    \caption{Ablation study for different quantization parameters. We only quantize one layer (Layer1.1.Conv2 or Layer4.1.Conv2) on RetinaNet-ResNet18 at W3A3. \#Params refers to the number of parameters to optimize.}
    \label{tab:tab_5}
\end{table}

\subsubsection{The Quantization Parameters in ODOL}

We only set the activation scale factor ($s_a$) as quantization parameters to optimize Eq~\ref{eq:local_reconstruction_ODOL} when given different $p$ values, then select the lowest $\mathcal{L}_{\text{ODOL}}$ and its corresponding $p$ value. However, for advanced PTQ, there is another kind of weight parameter named the rounding variable ($V$), which determines whether the weight values round up or down. As shown in Table \ref{tab:tab_5}, we experiment with optimizing $s_a$ and $V$ jointly as quantization parameters and find the optimal $p$ value of $L_p$ metric. The experiments show that $V$ is not sensitive to the optimization of optimal $p$ value. Introducing $V$ to the optimization will greatly increase the optimizing time, so we do not consider $V$. As to the weight scale factors, we follow the implementation of Qdrop and BRECQ with the optimization by minimizing MSE distance. 

\subsubsection{Effect of ODOL on Different Functions}
To show the effectiveness of different functions for the Object Detection Output Loss, we quantize RetinaNet-ResnetNet18 and Faster RCNN-ResNet18 at W4A4 as examples to conduct the ablation study. Table \ref{tab:tab_4} shows the experimental results. KL and $L_2$ measure the distance between the quantized and FP feature map for the classification head. IOU loss measures the spatial difference of bounding boxes from quantized and FP output. $L_1$ is leveraged for the difference of coordinate distance. We observe that choosing KL+$L_1$ as ODOL yields the best quantization result among all metrics. 
Hence we finally set KL+$L_1$ as the ODOL functions to represent the task information of object detection to guide the optimization of the adaptive $L_p$ metric.

KL divergence is widely used to measure the error of probability distribution. So it can better reflect the error of classification features than MSE. 
We also postulate that minimizing the difference of coordinate distance ($L_1$) directly rather than spatial difference (IOU loss) can better represent localization error for PTQ.
\begin{table}[tb]

\centering
\label{ablation}
\resizebox{0.80\columnwidth}{!}
{ 
\begin{tabular}{@{}cccccr@{}}
\toprule
\multirow{2}{*}{\textbf{Model}} & \multicolumn{4}{c}{\textbf{Metric}} &\multirow{2}{*}{\textbf{mAP}}\\ 
\cmidrule{2-5}
& ${L_2}$ & ${KL}$ & ${L_1}$ & ${IOU}$ &  \\ 
\midrule
 & $\checkmark$ & & $\checkmark$ &  & 31.07\\
\multirow{2}{*}{RetinaNet-ResNet18} &  & $\checkmark$ & $\checkmark$ &  & \textbf{31.12}\\
 & $\checkmark$ &  &  & $\checkmark$ & 30.21\\
 &  & $\checkmark$ &  & $\checkmark$ & 30.64\\
\midrule
 & $\checkmark$ & & $\checkmark$ &  & 33.41\\
\multirow{2}{*}{Faster RCNN-ResNet18} &  & $\checkmark$ & $\checkmark$ &  & \textbf{33.82}\\
 & $\checkmark$ &  &  & $\checkmark$ & 33.04\\
 &  & $\checkmark$ &  & $\checkmark$ & 33.11\\
\bottomrule
\end{tabular}
}
\caption{Ablation study of different combination for ODOL. We mark a $\checkmark$ when adopt the loss function. }
\label{tab:tab_4}
\end{table}
\subsubsection{Comparison with other Metrics}
To illustrate the superiority of our ODOL-based $L_p$ metric for a broader comparison, we conduct further experiments to compare it with other metrics.
In Table \ref{tab:tab_6}, we compare the adaptive BN-based $L_p$ and local MSE with our ODOL-based $L_p$ on the advanced PTQ method. Minimizing the distance of intermediate feature maps before and after quantization, those two metrics are denoted as the local metric. Our \modelname introduces the global performance loss information to determine the $L_p$ metric.
We can see that the ODOL-based adaptive $L_p$ metric can improve the accuracy by 0.58$\%$ compared with the MSE at the W4A4 bit setting. The ODOL also outperforms the BN to guide the $p$ value and improve by 2.36$\%$ on RetinaNet-ResNet18 compared with based on BN at W4A4. 

\begin{table}[htb!]

    \centering
    \resizebox{0.85\columnwidth}{!}
    { 
    
    \begin{tabular}{ccrr}
    
    \toprule
   \multirow{2}{*}{\textbf{Method}} & \multirow{2}{*}{\textbf{Metric}} & \multicolumn{2}{c}{\textbf{Bits(W/A)}}\\
   \cmidrule{3-4}
   & & W4A4 & W3A3\\
    \midrule
    Local & MSE & 30.54 & 28.48\\
      Local& BN-based adaptive $L_p$ & 28.76 & 27.16\\
     \textbf{Global}& \textbf{ODOL-based adaptive $L_p$} & \textbf{31.12} & \textbf{29.30}\\
      
    \bottomrule
    \end{tabular} 
    }
    \caption{Results of the parameter optimization by different metrics on RetinaNet-ResNet18 at W4A4.}
    \label{tab:tab_6}
\end{table}

\section{Conclusion}
In this paper, we explored the application of post-training quantization on object detection networks.
At first, we observed that the parameter $p$ in $L_p$ metric has a significant influence on quantization for object detection networks. 
And we indicated that it is necessary to set an adaptive $L_p$ metric.
To solve these problems, we propose the Object Detection Output Loss (ODOL) as an approximation of performance loss and then optimize the $p$ value using ODOL as the reconstruction quantization loss.
We proposed the framework, \modelname, which achieved the SOTA on networks for both 2D and 3D object detection tasks.

{\small
\bibliographystyle{ieee_fullname}
\bibliography{egbib}
}

\end{document}